# An argumentative annotation schema for meeting discussions

Vincenzo PALLOTTA, Hatem GHORBEL, Patrick RUCH and Giovanni CORAY

Faculty of Information and Communication Sciences
Swiss Federal Institute of Technology - Lausanne
IN F Ecublens 1015 Lausanne, Switzerland
{Vincenzo.Pallotta, Hatem.Ghorbel, Patrick.Ruch, Giovanni.Coray}@epfl.ch

**Abstract**

In this article, we are interested in the annotation of transcriptions of human-human dialogue taken from meeting records. We first propose a meeting content model where conversational acts are interpreted with respect to their argumentative force and their role in building the argumentative structure of the meeting discussion. Argumentation in dialogue describes the way participants take part in the discussion and argue their standpoints. Then, we propose an annotation scheme based on such an argumentative dialogue model as well as the evaluation of its adequacy. The obtained higher-level semantic annotations are exploited in the conceptual indexing of the information contained in meeting discussions.

## Introduction

Interaction through meetings is among the richest human communication activities. Multimodal multi-party dialogs can be audio-visually recorded and stored in a multimedia repository. Recording meetings implies the storage and the structuring of a large set of heterogeneous information scattered over time and media. The raw data format from the various recording devices is not directly usable for the creation of indexes, or for the content-based access to the relevant parts of the meeting recording. Data are then analysed and annotated in order to provide thematic access to the meeting recordings.

The application scenario we envisage for meeting recording, structuring, storage and retrieval is the following: suppose someone has not attended a group meeting, but needs information about "what happened" at the meeting. In this situation the user might want to make queries about the meeting participants, about the issues that were discussed and the decisions that were made. Answers to these queries can be of different types. An answer to the former question can be a list of participants, whereas a written summary or an excerpt of the most relevant audio-video recording sequence can be the answer to the latter. The user might also be interested in accessing the documents related to the meeting, such as the agenda, reports, presentation handouts, related articles, etc.

In the framework of the (IM)2[1] project, we are aimed at developing a robust computational dialogue model which could be exploited for the representation and the access to the information contained in dialogues from meeting discussions. In this article, we propose an annotation schema that describes the task of argumentative annotation of meeting discussions.

The goal of such an annotation is to enrich the structure of meetings by means of higher-level semantic description.

Based on such a description, we also propose to apply Information Retrieval (IR) and Question Answering (QA) techniques to create semantic indexes that aid in the answering content-based queries about the meetings.

## Meeting annotation

A first type of annotation is based on the shallow dialogue model, proposed in (Armstrong et al., 2003). This model provides a simple logical structure for dialogues based on the following categories:

- a *dialog* is a non empty set of episodes; a new episode is identified by a topic shift.
- an *episode* is a non empty set of turns; a new turn is introduced by a speaker change.
- a *turn* is a non empty sequence of utterances.

Each turn is annotated by one or more dialogue acts highlighting the communicative function of an utterance. The set of dialogue act labels is based on Switchboard/DAMSL guidelines (Core and Allen, 1997) and it is currently used for the annotation of the ICSI[2] corpus of meeting dialogues that we use for our tests.

In addition to the shallow model, which will be mostly automatically extracted, we also consider the adoption of a more structured representation. It will be produced with more manual intervention, and will provide the system with more possibilities for the users to identify the desired segments of the meeting. In fact, the main limitation of the shallow dialogue model is that a single utterance may have multiple communicative functions and that there is no trace of participants' intentions. In multimodal dialogues, for instance, there are other types of communicative actions besides utterances, e.g. agreement by applause, disagreement by gesture or facial expressions. Also silence might express an agreement after a question like "Do you have something to object?". Moreover, the model does not take into account the emergence of opinions by hearers about speakers, and, more important, it does not highlight the social behaviour of the participants, nor their role in the deliberation process.

In order to overcome the above limitation, we propose to consider meeting dialogues from the Collaborative Decision Making (CDM) perspective (Pallotta, 2003). A meeting is defined as a multi-party (multi-agent) decision making process: a collaborative process, where agents follow a series of communicative actions in order to establish a common ground on the dimension of the problem. The main dimensions of CDM process are:

---

[1] The National Center of Competence in Research (NCCR) on Interactive Multimodal Information Management, in brief (IM)2, http://www.im2.ch, is aimed at the advancement of research, and the development of prototypes, in the field of man-machine interaction.

[2] International Computer Science Institute, Berkeley. http://www.icsi.berkeley.edu

- An overall task goal;
- A set of alternatives;
- A collection of choice criteria (perspectives and preferences) settled upon the participants;
- A decision (or evaluation) function that combines criteria to judge the alternatives.

This definition focuses on the processes, which take place during meetings and how these processes contribute to the accomplishment of a joint goal.

**Types of Meetings**

According to a classification provided by NIST (Cugini et al, 1997) and based on the McGrath work on group dynamics (McGrath, 1984), meeting scenarios are likely to contain the following CDM processes:

***Staff Meetings:*** Participants discuss real technical is-sues, brainstorms ideas and make decisions. They include also planning, negotiation and brainstorming.

***Information exchange and decision-making meetings:*** For instance, office furnishing: An expert will help participants to choose office furniture, carpet, etc. for an office. They also include brainstorming.

***Information gathering and decision-making meetings:*** For instance, shopping on-line where participants search the Web and collaborate to purchase a digital camera. This type of meeting includes also negotiation.

As resulted from recent studies (Pallotta et al., 2004), questions related to the above type of meetings are mainly those pertaining to the outcome of the discussion in terms of the arguments invoked, questions raised, and consensus achieved on the discussed issues. There are few examples of possible question types:

*About discussion:*
1. What were the objections against the proposal Y?
2. What was the position of X on subject Z?
3. Give me all the contributions of participant X in favour of alternative A regarding the issue I.
4. Who was supporting the alternatives proposed by X?
5. Who systematically rejected all the proposals made by X?

*About the decision:*
6. Which alternatives have been chosen for the issue I?
7. Why the alternative A has been rejected for the issue I?
8. For which open question there was no solution adopted? Why? What are the open questions for a next meeting?
9. Which criteria were chosen to take the decision D1?
10. Which criteria did the members who disagreed on the decision D1 invoke?

*About the coherence of the dialogue:*
11. When did X contradict himself about the issue I?
12. Was the decision about issue X democratically taken?

## Argumentative structure of meetings

In order to answer the types of questions exemplified in the previous section, we need to further mark up meeting recordings with appropriate meta-description. In other words, we need to annotate the parts of the meeting where the decision process takes place with a suitable "argumentative structure".

A basis for an Argumentation Mark-Up language has been proposed in (Delannoy, 1999). It provides a set of XML tags to pre-process text containing arguments in order to build summaries. We believe that this model is not sufficient for the meeting dialogues, since it only highlights argumentative rhetorical relations in monologues.

Figure 1: The IBIS model

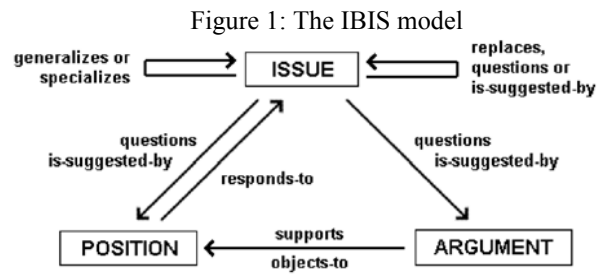

A simple model, shown in figure 1, of an argumentative structure is the "Issue Based Information Systems" (IBIS), proposed by (Kunz and Rittel, 1970) and adopted as a foundational theory in some computer-supported collaborative argumentation (CSCA) systems such as Zeno (Gordon and Karacapilidis, 1999), HERMES (Karacapilidis and Papadias, 2001), Questmap (Conklin et al., 2001), and Compendium (Selvin, 2001). We adopt this model as the reference model for the description of the argumentative structure of decision meetings. The model captures and highlights the essential lines of a discussion in terms of what issues have been discussed and what alternatives have been proposed and accepted by the participants.

The development of the argumentation structure is a dynamic process which itself needs to be modelled. In (Gordon and Karacapilidis, 1999), the process of proposing and arguing over alternatives is modelled by a state transition graph. This simple model uses a strict protocol, which constrains the interaction between participants, but hardly scales up from computer-mediated discussion to real life, unconstrained, meetings.

One main difference between CSCA and meeting recordings is that, in the former case, the argumentative structure is used to constrain the interaction whereas, in the latter, we aim at deriving the corresponding argumentative structure (if any) by observing a recorded interaction. It is apparent that the second problem is harder since one has to infer the causal relationships of dialogue events without knowing the participants' intentions and goals. Modelling the dynamics of argumentation also means dealing with multimodal knowledge about the dialogue events, since some argumentation acts can be stated in other modalities (e.g. agreement by silence or by applause, disagreement by laughing).

The importance of tracking collaborative argumentation of discussion meetings has a central importance for building what Duska Rosemberg calls "Project Memories" in (Rosemberg and Sillince, 1999). The construction of projects memories is similar to the annotation of meetings by their argumentation structure since it highlights not only "strictly factual, technical information", but also relevant information about the decision making process.

**The Meeting Description Schema**

The description schema we propose as the starting point for the construction of a *meeting ontology*, is formalized using XML-schemas[3] and reflects the substantial aspects of the IBIS model. However, we only model argumenta-tive relationships between meeting episodes, without imposing any strong constraint on the structure that the meeting develops. The Meeting Description Schemas (MDS) is based on the previous observation that there exist a number of sequencing regularities in dialogue, adjacency pairs, de-scribing facts as, for instance, that questions are generally followed by answers, issues by solutions, proposals by acceptances or rejections, etc. The main point in our proposal is that adjacency pairs are described within a specific dialogue context where they may have different argumentative roles and thus different interpretations. For instance, accepting a proposal in a context of the agenda discussion in a meeting has different argumentative role than accepting a proposal in a context of an issue discussion. The context is basically seen as a temporal interval during which the discussion has a specific focus: an episode. In MDS, the dialogue contexts are represented by episodes and can be viewed as snap-shots of the discussion during which a specific argumentation act is occurring. As argumentation acts could be decomposed in more specific or detailed sub-acts, an episode itself can be com-posed of other sub-episodes. For instance an episode of agenda in a meeting can be detailed into an episode of agenda presentation, some episodes of propositions to add (or modify or delete) an issue to be discussed in the agenda, as shown in figure 2.

---
[3] http://www.w3.org/XML/Schema/

Figure 2: Meeting's main structure

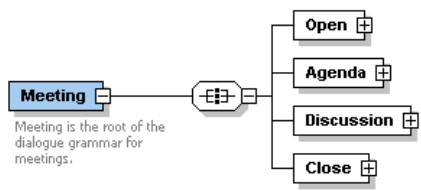

Although episodes could overlap in time, we can represent the taxonomy of the Meeting Schema as a hierarchical structure: each argumentative act is composed of possible sub-acts, as shown in figure 3 for the case of the Agenda episode.

Figure 3: Agenda episode structure

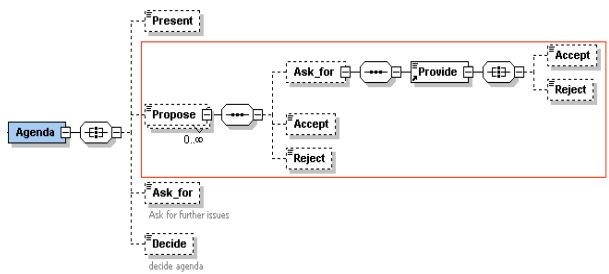

However, practically, when analyzing the dialogue, the hierarchical structure is not sufficient to represent the adjacency pairs: consider an answer that refers to two questions in the discussion. In this case, we need to add a relation that links the answer to both of the questions. This relation is called "reply-to" relation, which links an episode to one or more previous (in time) episodes. The reply-to relation induces a chain structure on the dialogue which is local to each episode and which enables to visu-alise its context. For instance, the context of *ACCEPT(clarification)* will be the episode of the clarification and that of the proposal (if we know that a clarification is preceded by a proposal) as well as the episode where the proposal was uttered (agenda, discussion issue, etc.).

Categories such as ASK, ACCEPT, REJECT might (but not necessary) correspond to dialogue acts. In this case we have refined the concept of dialogue act and adjacency pairs by specifying the role of dialogue contribution within the discussion.

Note that there is an invariant structure of discussion episodes, which is present in several discussion episodes (changing the external parameter i.e. "discuss-issue"), which is reported below and framed in figure 3 for the case of Agenda; This structure mirrors in terms of episodic structure, the IBIS model. The only structural constraints are imposed to the "reply-to" relation, which is graphically represented above by dependency tree. In fact, we require that an argumentative episode "replies" to the parent argumentative episode in the argumentation tree:

*reply-to(ACCEPT(explanation),PROVIDE(explanation)).*

If considered as constraints, this model can be viewed as a sort of argumentation dependency grammar.

Finally, from an annotator's point of view, more general categories can be refined by sub-categories if the annotator is able to identify finer-grained episodes. This means that the annotation can be done in several stages, following either a top-down (from general to specific categories) or bottom-up strategy

The automatic construction of argumentative structures from meetings is a long-term goal, while manual construction seems to be more reasonable in the short-term. Nevertheless, there are some important steps, which can be performed automatically. Recent works (McCowan et al., 2002) made by partners in the (IM)2 project have been focused on stochastic models for the automatic detection of meeting episodes based on the combined extraction of multiple audio-visual features from meeting recordings. Indeed, the automatic detection of candidate segments where an argumentation is likely to be found could be of great help to the human annotator. Moreover, in an adaptive annotation tool (Ballim et al., 2000), topic segmentation can be used to propose annotation of issues, and the presence of a high number of propose-accept or propose-reject adjacency pairs may signal a segment, which contains an argumentation act.

## Towards a Meeting Query Engine

Searching meeting dialogues raises several problems compared to standard IR indexing techniques. One important point is that users may ask different types of queries depending on their needs, therefore one single retrieval strategy may be not sufficient. The required recall and precision may vary depending on the type of query since, in some cases the user's query needs to be answered with a precise information (e.g. who, where, when, why), while in some other cases the user is looking for a context (i.e. a passage in the meeting) where the interesting information can be found. Moreover, it is not apparent which level of granularity the document base should have (turns, episodes, issues or entire meetings).

If we also aim at navigation through meetings by query, the problem of contextual interpretation of user's queries becomes more evident (e.g. history-based IR). The link to additional knowledge (present in the meeting repository in form of annotations or reference to other knowledge sources or related documents) may increase the robustness and the performance of the search engine (e.g. imagine the situation where the user is asking for a passage where a topic is only contained in the slide presentation which a meeting participant is referring to during the meeting, without explicitly mentioning the topic in the dialogue). IR techniques can be applied to the following problems:

- *inter-meeting* search: look for relevant meetings among a collection of meetings;
- *intra-meeting* search: search within a single meeting for relevant (multimodal/multimedia) passage (topics, episodes, documents, ..);
- *entity* search: look for specific entities (e.g. participants, issues, documents, dates) within a single or multiple meetings (e.g. "who was the person who systematically disagree with X's proposals?");
- *discussion's event* search: look for certain action/events which have a particular role in the discussion (e.g. proposals, argument, counter-argument, support, rejection, acceptance).

We believe that standard text-based IR techniques are adequate to meet the requirements for retrieving multi-party multimodal dialogues only partially. In a real application we suspect that the users will ask for complex and focused queries that could not be adequately answered (in terms of both precision and recall) if the indexing model does not include some information sources other than textual data. In practice we propose to extend the traditional IR techniques to dialogues by combining heterogeneous indexes having different nature (lexical, semantic) and using different modalities (speech, documents). Each dialogue segment (episode, turn, utterance) will thus be referenced in four indexes:

- a stem-based index;
- an automatically-built[4] latent argumentative index (PURPOSE, METHODS, CONCLUSION) (Ruch et al., 2003);
- a human (or semi automatically) annotated argumentative index described in MDS (PROPOSE, ISSUE, ACCEPT, REJECT, DECISION, …);
- a speech to document dynamic index.

One might consider, in addition to enhanced content indexing, investigating on the knowledge-based methods for flexible query expansion/reformulation.

## Conclusion

We believe that a Query Engine on records of meeting discussions requires adequate conceptual indexing techniques. We propose an annotation scheme based on a model of argumentation. The scheme provides the right level of abstraction and meets the requirements of the targeted application (Pallotta et al. 2003), since the argumentative structure is needed to answer actual queries about issues and events that occur in meeting discussions.

## References

Armstrong, S., Clark, A., Coray, G., Georgescul, M., Pallotta, V., Popescu-Belis, A., Portabella, D., Rajman, M., and Starlander, M. (2003). Natural Language Queries on Natural Language Data: a Database of Meeting Dialogues. In Proceedings of NLDB 2003, Burg/Cottbus, Germany, 2003.

Ballim, A., Pallotta, V., Fatemi, N., and Ghorbel, H. (2000). A knowledge-based approach to semi-automatic annotation of multimedia documents via user adaption. In proceedings of the EAGLES/ISLE Workshop on Meta-Descriptions and Annotation Schemas for Multimodal/Multimedia Language Resources (LREC 2000 Pre-Conference Workshop), Athens, Greece.

Conklin, J., Selvin, A., Buckingham Shum, S., and Sierhuis, M. (2001). Facilitated Hypertext for Collective Sensemaking: 15 years on from gIBIS: Knowledge Media Institute, Open University.

Core, M, and Allen, J. (1997). Coding Dialogs with the DAMSL Annotation Scheme. Paper presented at AAAI Fall Symposium on Communicative Action in Humans and Machines, Boston, MA.

Delannoy, J-F. (1999). Argumentation Mark-Up. Paper presented at Walker M. (ed) proceedings of ACL Workshop on "Towards standards and tool for discourse tagging".

Cugini, J. et al. (1997). Methodology for Evaluation of Collaborative Systems. The Evaluation Working Group of the DARPA Intelligent Collaboration and Visualization Program, Revision 3.0: August, 27, 1997.

Gordon, T., and Karacapilidis, N. (1999). The Zeno Argumenta-tion Framework. German Artificial Intelligence 3:20-29.

Karacapilidis, N., and Papadias, D. (2001). Computer Supported Argumentation and Collaborative Decision Making: The HERMES system. Information Systems 26, no 4:259-277.

Kunz, W., and Rittel, H. W. J. (1970). Issues as elements of information systems: Universität Stuttgart, Institüt für Grundlagen der Planung.

McCowan, I., Bengio, S., Gatica-Perez, D., Lathoud, G., Monay, F., Moore, D., Wellner, P., and Bourlard, H. 2002. Modeling Human Interaction in Meetings. Martigny: IDIAP.

McGrath, J. E. (1984). Groups: Interaction and Performance. Englevood Cliffs, N. J., Prentice-Hall.

Pallotta V., A computational dialectics approach to meeting tracking and understanding, in Giacalone-Ramat, Rigotti, Rocci (eds.) Special issue on Linguistics and new professions, Materiali Linguistici , Franco Angeli, October 2003.

Pallotta V., Lisowska A. and Marchand-Maillet S. Towards Meting Information Systems: MeetingKnowledge Management. In proceeding of ICEIS 2004 international conference, 14-17 April, 2004, Porto, Portugal.

Ruch, P., Chichester, C., Cohen, G., Coray, G., Ehrler, F., Ghorbel, H., Müller, H, Pallotta, V. Report on the TREC 2003 Experiment: Genomic Track, TREC 2003, TREC Proceedings, Gaithersburg, 17-21 November 2003.

Rosemberg, D., and Sillince, J. A.A. 1999. Common Ground in Computer-Supported Collaborative Argumentation. Paper presented at Workshop on Computer Supported Collaborative Argumentation for Learning Communities (CSCL99).

Selvin, A. et al. (2001). Compendium: Making Meetings into Knowledge Events. In Proceedings of Knowledge Technologies 2001 March 4-7, Austin TX.

---

[4] For further details look at Latent Argumentative Structure (LASt) tool, http://lithwww.epfl.ch/~ruch/softs/softs.html